\documentclass[lettersize,journal]{IEEEtran}
\usepackage{amsmath,amsfonts}
\usepackage{algorithmic}
\usepackage{algorithm}
\usepackage{array}
\usepackage[caption=false,font=normalsize,labelfont=sf,textfont=sf]{subfig}
\usepackage{textcomp}
\usepackage{stfloats}
\usepackage{url}
\usepackage{verbatim}
\usepackage{graphicx}
\usepackage{cite}
\begin{document}

\title{Stepwise Model Reconstruction of Robotic Manipulator Based on Data-Driven Method}

\author{Dingxu Guo, Jian Xu, Shu Zhang
\thanks{D. Guo, J. Xu and S. Zhang are with the School of Aerospace Engineering
and Applied Mechanics, Tongji University, Shanghai, China.(Corresponding
author e-mail: zhangshu@tongji.edu.cn)}
\thanks{}}

\markboth{}%
{Shell \MakeLowercase{\textit{et al.}}: A Sample Article Using IEEEtran.cls for IEEE Journals}

\IEEEpubid{}

\maketitle

\begin{abstract}
Research on dynamics of robotic manipulators provides promising support for model-based control. In general, rigorous first-principles-based dynamics modeling and accurate identification of mechanism parameters are critical to achieving high precision in model-based control, while data-driven model reconstruction provides alternative approaches of the above process. Taking the level of activation of data as an indicator, this paper classifies the collected robotic manipulator data by means of K-means clustering algorithm. With the fundamental prior knowledge, we find the corresponding dynamical properties behind the classified data separately. Afterwards, the sparse identification of nonlinear dynamics (SINDy) method is used to reconstruct the dynamics model of the robotic manipulator step by step according to the activation level of the classified data. The simulation results show that the proposed method not only reduces the complexity of the basis function library, enabling the application of SINDy method to multi-degree-of-freedom robotic manipulators, but also decreases the influence of data noise on the regression results. Finally, the dynamic control based on the reconfigured model is deployed on the experimental platform, and the experimental results prove the effectiveness of the proposed method.
\end{abstract}


\section{Introduction}
\IEEEPARstart{A}{t} present, robotic manipulators have been widely applied in different industrial fields, and the demand for high speed and high precision in industry has placed tough requirements on the control performance of robotic manipulators. Previous studies have shown that dynamics research of robotic manipulators can effectively guide the design of model-based control laws\cite{ref1} and thus improve control performance\cite{ref2}. A reliable dynamics model can take advantage of model-based control methods, while an inaccurate model can even be counterproductive, which increases the threshold of application of model-based control methods for industrial robotic manipulators.

According to the traditional approach, a reliable dynamics model of a robotic manipulator is obtained in two main steps: dynamics modeling by first principles and parameter identification based on the developed model. For multi-degree-of-freedom robotic manipulators, the first principles modeling is a complex process that requires significant expertise, and usually requires simplified assumptions and idealized situations, while some factors exist in real robotic manipulators that are difficult to model, such as joint friction and nonlinear characteristics of components, which typically result in modeling errors. In addition, the reliable dynamics model requires numerous parameters, such as mass, rotational inertia, and friction coefficient, that often need to be identified. However, parameter identification is based on the developed model, and factors impacting the accuracy of the modeling still affect the identification results, thus limiting the accuracy of the model. In contrast, data-driven modeling approaches use collected data to infer the dynamic behavior of the system, which avoid the complex mathematical derivation and modeling process, as well as the modeling errors caused by idealized assumptions, and have been widely used and developed in practical applications.

Data-driven modeling approaches are divided into two categories in accordance with model form, i.e., machine learning methods with black-box models and symbolic regression methods with white-box analytical models. The former has powerful fitting capabilities to handle complex nonlinear relationships in dynamics of robotic manipulators. However, the machine learning models characteristic by black-box forms do not provide a physical perspective on the established relationships\cite{ref3}. In contrast, the latter uses the data and regresses to obtain an analytic form model, so that further theoretical analysis and some advanced model-based control methods can be achieved. Genetic Programming (GP), as a symbolic optimization technique, enables the regression of symbolic structures for nonlinear systems from data\cite{ref4}. Currently, GP-based symbolic regression methods have been successfully applied in various cases, including finding closed-form solutions for inverse kinematics of robotic manipulator\cite{ref5}, reconstructing the dynamics model of multi-degree-of-freedom robotic manipulators \cite{ref6}, and guiding controller design\cite{ref7}. However, the gradient-free learning process of GP poses a challenge in striking a balance between symbolic complexity and model performance during the symbolic regression, often resulting in code bloat. Additionally, for high-dimensional nonlinear systems of multi-degree-of-freedom robotic manipulators, the symbolic forms obtained through GP regression tend to be excessively complex(i.e., $\sin \left( {\sin \left( {\sin x} \right)} \right)$), making it difficult to discern the underlying physical interpretations of the symbols.

As a powerful complement, sparse identification of nonlinear dynamics (SINDy) proposed by Steven L. Brunton\cite{ref8} is also an effective symbolic regression method of finding underlying dynamics of nonlinear systems such as Lorentz system. Afterwards, sparse identification has been applied in finding the variational law\cite{ref9} and the Lagrangian from learned energy\cite{ref10} hidden in physical systems, reconstructing Jacobian matrix\cite{ref11} and gravity term\cite{ref12} of robotic manipulators. The key aspect of the SINDy lies in incorporating prior knowledge to construct a suitable library of nonlinear functions that adequately describes the dynamic behavior of the system, so that SINDy achieves favorable results when the symbolic form of the target model is relatively concise. However, SINDy encounters the challenge of an excessively large library of nonlinear functions when dealing with tasks such as reconstructing complex dynamical models, i.e., multi-degree-of-freedom robotic manipulators. Scholars have proposed optimization methods such as SINDy-PI\cite{ref13} for this problem, but this limitation still further restricts SINDy's generalizability to complex systems.

Clustering, as an unsupervised learning method that enables grouping from data based on similarity or correlation, has been widely used in the fields of data analysis, pattern recognition and information processing\cite{ref14}. In recent years, researchers have also recognized the potential of clustering algorithms in the applications of symbolic regression and model reconstruction. For instance, Daniel Fernex\cite{ref15} proposed the cluster-based network modeling method for data-driven modeling of complex nonlinear dynamics from time-resolved snapshot data, as well as Jason J\cite{ref16} found the slow timescale characteristics of multiple timescales system by using clustering techniques and extracted the slow timescale dynamics using SINDy. If data can be classified based on their features using unsupervised clustering techniques, it would reduce the complexity of the dynamical systems and enable the extension of symbolic regression methods to even more complex systems.

The remainder of this paper is organized as follows. In Section II, the SINDy  based dynamic model reconstruction of robotic manipulators is briefly introduced and its limitations are analysed, and then activation is used as an indicator for the clustering analysis of the collected data. In Section III, a stepwise model reconstruction method, i.e., stepwise SINDy, is proposed based on the clustering results, and a 3-DoF simulation platform is used to demonstrate the advantages of stepwise SINDy in terms of sparsity and extrapolated prediction accuracy in the cases with data noise. Section IV introduces a three-axis experimental platform to validate the effectiveness of the proposed stepwise model reconstruction method. The conclusion of this paper is described in Section V.

\section{Data Clustering}
\subsection{Revisit of SINDy}
Dynamics model of robotic manipulator, relating robot motion to joint torques, plays an important role in the design of advanced control laws. For a $n$-degree-of-freedom ($n$-DoF) serial manipulator composed of rigid links and rotational joints, the dynamical equations can be derived through first principles, such as the Lagrangian method. Ignoring the joint friction and the external forces, the dynamics model of the system can be generally written in the following form
\begin{equation}
\label{eq_1}
{\mathbf{M}}\left( {\mathbf{q}} \right){\mathbf{\ddot q}} + {\mathbf{C}}\left( {{\mathbf{q}},{\mathbf{\dot q}}} \right){\mathbf{\dot q}} + {\mathbf{G}}\left( {\mathbf{q}} \right) = {\mathbf{\tau }},
\end{equation}
where $\mathbf{q}$ is the joint angles of the manipulator and $\mathbf{M}$, $\mathbf{C}$, $\mathbf{G}$ respectively represent the mass matrix, the Coriolis and centrifugal matrix and the gravity term, $\mathbf{\tau}$ is the vector of joint driving torques. According to the concept of SINDy, a function can be sparsely represented as a linear combination of nonlinear terms. Denote by $\mathbf{x}(t)=\left[\begin{array}{ll}\mathbf{q} & \dot{\mathbf{q}}\end{array}\right]^{\mathrm{T}}$ the state of the system at a time  , the dynamics model can also be expressed as
\begin{equation}
\label{eq_2}
{\frac{d}{{dt}}{\mathbf{x}}\left( t \right) = {\mathbf{f}}\left( {{\mathbf{x}}\left( t \right)} \right).}
\end{equation}
However, the explicit expression for the angular acceleration $\mathbf{\ddot q}$ of the robotic manipulator is difficult to obtain, as well as there is a greater focus on the relationship between the robot motion and joint torques. Each driving torque can be formulated by
\begin{equation}
\label{eq_3}
{{\tau _i} = {\mathbf{h}}\left( {{\mathbf{q}},{\mathbf{\dot q}},{\mathbf{\ddot q}}} \right) = {\mathbf{Y}}\left( {{\mathbf{q}},{\mathbf{\dot q}},{\mathbf{\ddot q}}} \right){{\mathbf{\Xi }}_i},{\text{ }}i = 1,2, \ldots ,n,}
\end{equation}
where ${\mathbf{Y}}\left( {{\mathbf{q}},{\mathbf{\dot q}},{\mathbf{\ddot q}}} \right)$ is the library of candidate functions and ${{\mathbf{\Xi }}_i}$ is the sparse coefficients vector of $i$th joints. Due to the characteristics of rotational joints of the serial manipulator, ${\mathbf{Y}}\left( {{\mathbf{q}},{\mathbf{\dot q}},{\mathbf{\ddot q}}} \right)$ consists of constants, trigonometric functions of joint angles and polynomials. Then, driving the serial manipulator under trajectories with different motion patterns and recording the signal of robot motion and joint torques and collecting data, we have
\begin{equation}
\label{eq_4}
{\left[ {\begin{array}{*{20}{c}}
  {{\tau _i}\left( {{t_1}} \right)} \\ 
  {{\tau _i}\left( {{t_2}} \right)} \\ 
   \vdots  
\end{array}} \right] = \left[ {\begin{array}{*{20}{c}}
  {{Y_1}({t_1})}& \cdots &{{Y_p}({t_1})} \\ 
  {{Y_1}({t_2})}& \cdots &{{Y_p}({t_2})} \\ 
   \vdots  & \vdots & \vdots  
\end{array}} \right]\left[ {\begin{array}{*{20}{c}}
  {{\xi _{i1}}} \\ 
  {{\xi _{i2}}} \\ 
   \vdots  \\ 
  {{\xi _{ip}}} 
\end{array}} \right].}
\end{equation}
At last, the sparse coefficients vector ${{\mathbf{\Xi }}_i}$ is determined by solving the regression problem, i.e., LASSO regression
\begin{equation}
\label{eq_5}
{\mathop {\min }\limits_{{{\mathbf{\Xi }}_i}} {\left\| {{{\mathbf{\tau }}_i} - {\mathbf{Y}}{{\mathbf{\Xi }}_i}} \right\|_2} + \lambda {\left\| {{{\mathbf{\Xi }}_i}} \right\|_1},}
\end{equation}
where $\lambda$ is the sparsity promoting parameter.

\subsection{Data Clustering Based on Activation Index}

As the number of DoF of serial manipulator increases, the form of the candidate functions in ${\mathbf{Y}}\left( {{\mathbf{q}},{\mathbf{\dot q}},{\mathbf{\ddot q}}} \right)$ will grow more complex and lead to curse of dimensionality. Furthermore, it’s well known that the noise of velocity and acceleration in actual robot signal acquisition cannot be ignored. We need to reflect on whether data containing more information is more beneficial for identification or model reconstruction. Following this idea, an index about activation is designed as follows
\begin{equation}
\label{eq_6}
{I\left( {\mathbf{a}} \right) = S\left( {\mathbf{a}} \right) + W{\left\| {\mathbf{a}} \right\|_1}{\text{sign}}\left( {\sum\limits_{i = 1}^n {{a_i}} } \right),}
\end{equation}
where ${\mathbf{a}} = {\left[ {\begin{array}{*{20}{c}}
  {{a_1}}&{{a_2}}& \cdots &{{a_n}} 
\end{array}} \right]^{\text{T}}}$, $W$ is the weight coefficient of activation level, and $S(\bullet)$ is the function of activation switch that takes the form
\begin{equation}
\label{eq_7}
{S\left( {\mathbf{a}} \right) = \tanh \left( {h{{\left\| {\mathbf{a}} \right\|}_1}} \right) = \frac{{1 - {e^{ - h{{\left\| {\mathbf{a}} \right\|}_1}}}}}{{1 + {e^{ - h{{\left\| {\mathbf{a}} \right\|}_1}}}}},}
\end{equation}
where $h$ is the parameter that adjusts the sensitivity of the activation switch. After pre-processing the velocity and acceleration data using the activation index, the collected data can be clustered according to the classical k-means method. The procedure of the data clustering based on activation index is shown in Algorithm 1.

\begin{figure*}[!t]
\centering
\includegraphics[width=6.5in]{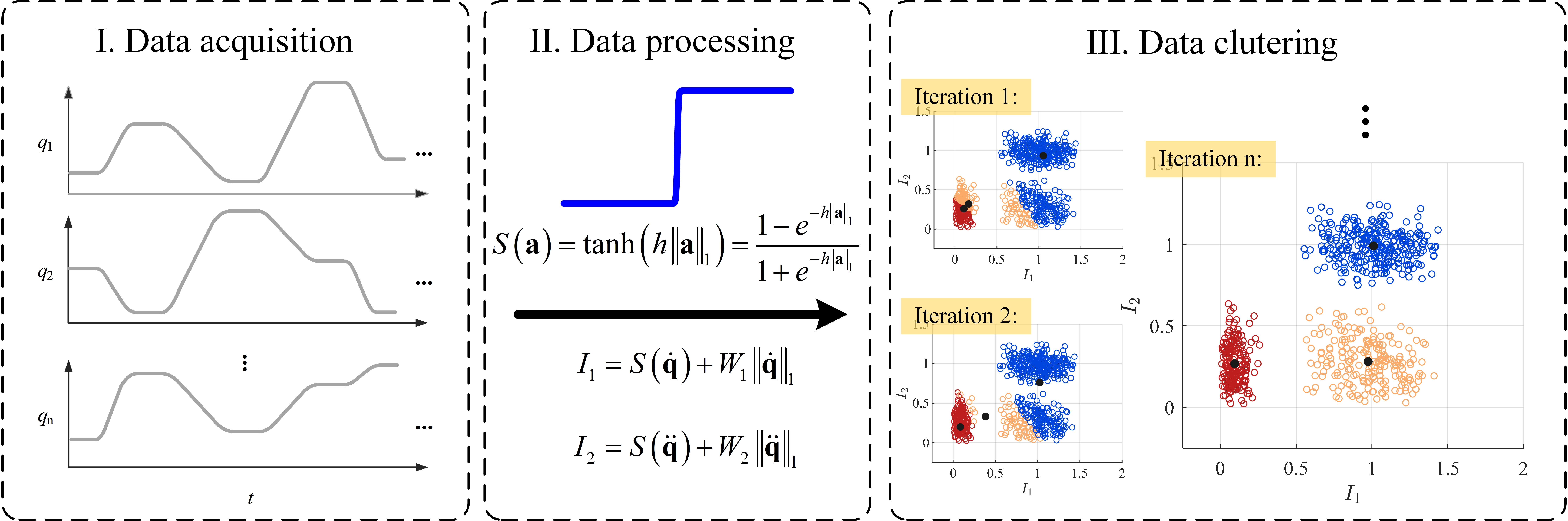}
\caption{The diagram of data clustering based on activation index. Data is collected in a simulation of $n$-DoF serial manipulator positioning and sorting conditions, and $k$-means method is used in clustering the data after activation index processing. Here the parameters are chosen as $h=60$, $W_1=0.6$ and $W_2=0.32$, respectively.}
\label{fig_1}
\end{figure*}

\begin{algorithm}[H]
\caption{Procedure of data clustering based on activation index.}\label{alg:alg1}
\begin{algorithmic}
\STATE 
\STATE $\textbf{Input:}\text{Collected \textit{m} sets of data}$
\STATE $\textbf{Data processing:}$
\STATE \hspace{0.5cm}$ \text{Give the value of parameters $W_1$, $W_2$ and   h}$
\STATE \hspace{0.5cm}$ \text{Calculate ${\mathbf{x}} = \left[ {\begin{array}{*{20}{c}}
  {I\left( {{\mathbf{\dot q}}} \right)}&{I\left( {{\mathbf{\ddot q}}} \right)} 
\end{array}} \right]$ for each data set}$
\STATE $\textbf{Data Clustering: k-means method}$
\STATE $\textbf{Input:} \text{$\left\{ {\begin{array}{*{20}{c}}
  {{{\mathbf{x}}_1}}&{{{\mathbf{x}}_2}}& \cdots &{{{\mathbf{x}}_m}} 
\end{array}} \right\}$, number of clusters $k$, and}$
\STATE \hspace{0.9cm}$\text{maximum iterations $\lambda_{max}$}$
\STATE \hspace{0.5cm}$ \text{Select randomly initial clustering centers}$
\STATE \hspace{0.5cm}$\left\{ {\begin{array}{*{20}{c}}
  {{{\mathbf{\mu }}_1}}&{{{\mathbf{\mu }}_2}}& \cdots &{{{\mathbf{\mu }}_k}} 
\end{array}} \right\}$
\STATE $\textbf{Repeat}$
\STATE \hspace{0.5cm}$ \text{Let ${\mathbb{C}_j} = \emptyset ,j = 1,2, \cdots ,k$}$
\STATE \hspace{0.5cm}$\textbf{for }\text{$i = 1,2, \cdots ,m$}\textbf{ do}$
\STATE \hspace{1cm}$\text{Calculate ${d_{ij}} = {\left\| {{{\mathbf{x}}_i} - {{\mathbf{\mu }}_j}} \right\|_2},j = 1,2, \cdots ,k$}$
\STATE \hspace{1cm}$\text{Let $\gamma  = \arg {\min _{j \in \left\{ {1,2, \cdots ,k} \right\}}}{d_{ij}}$, ${\mathbb{C}_\gamma } = {\mathbb{C}_\gamma } \cup \left\{ {{{\mathbf{x}}_i}} \right\}$} $
\STATE \hspace{0.5cm}$\textbf{end for}$
\STATE \hspace{0.5cm}$\text{Let $\lambda =0$}$
\STATE \hspace{0.5cm}$\textbf{for }\text{$j = 1,2, \cdots ,k$}\textbf{ do}$
\STATE \hspace{1cm}$\text{Denote $c_j$ as the number of elements of $\mathbb{C}_j$}$
\STATE \hspace{1cm}$\text{Calculate the gravity position of $\mathbb{C}_j$: ${{\mathbf{\mu '}}_j} = \frac{{\sum\nolimits_{{\mathbf{x}} \in {\mathbb{C}_j}} {\mathbf{x}} }}{{{c_j}}}$}$
\STATE \hspace{1cm}$\textbf{if ${{\mathbf{\mu '}}_j} \ne {{\mathbf{\mu }}_j}$} $
\STATE \hspace{1.5cm}${{\mathbf{\mu }}_j} = {{\mathbf{\mu '}}_j}$
\STATE \hspace{1cm}$\textbf{else}$
\STATE \hspace{1.5cm}$\lambda=\lambda+1$
\STATE \hspace{1cm}$\textbf{end if}$
\STATE \hspace{0.5cm}$\textbf{end for}$
\STATE $\textbf{until $\lambda=\lambda_{max}$}$
\STATE $\textbf{Output:}\text{Classified clusters $\left\{ {\begin{array}{*{20}{c}}
  {{\mathbb{C}_1}}&{{\mathbb{C}_2}}& \cdots &{{\mathbb{C}_k}} 
\end{array}} \right\}$}$
\end{algorithmic}
\label{alg1}
\end{algorithm}

\begin{figure}[!t]
\centering
\includegraphics[width=2.5in]{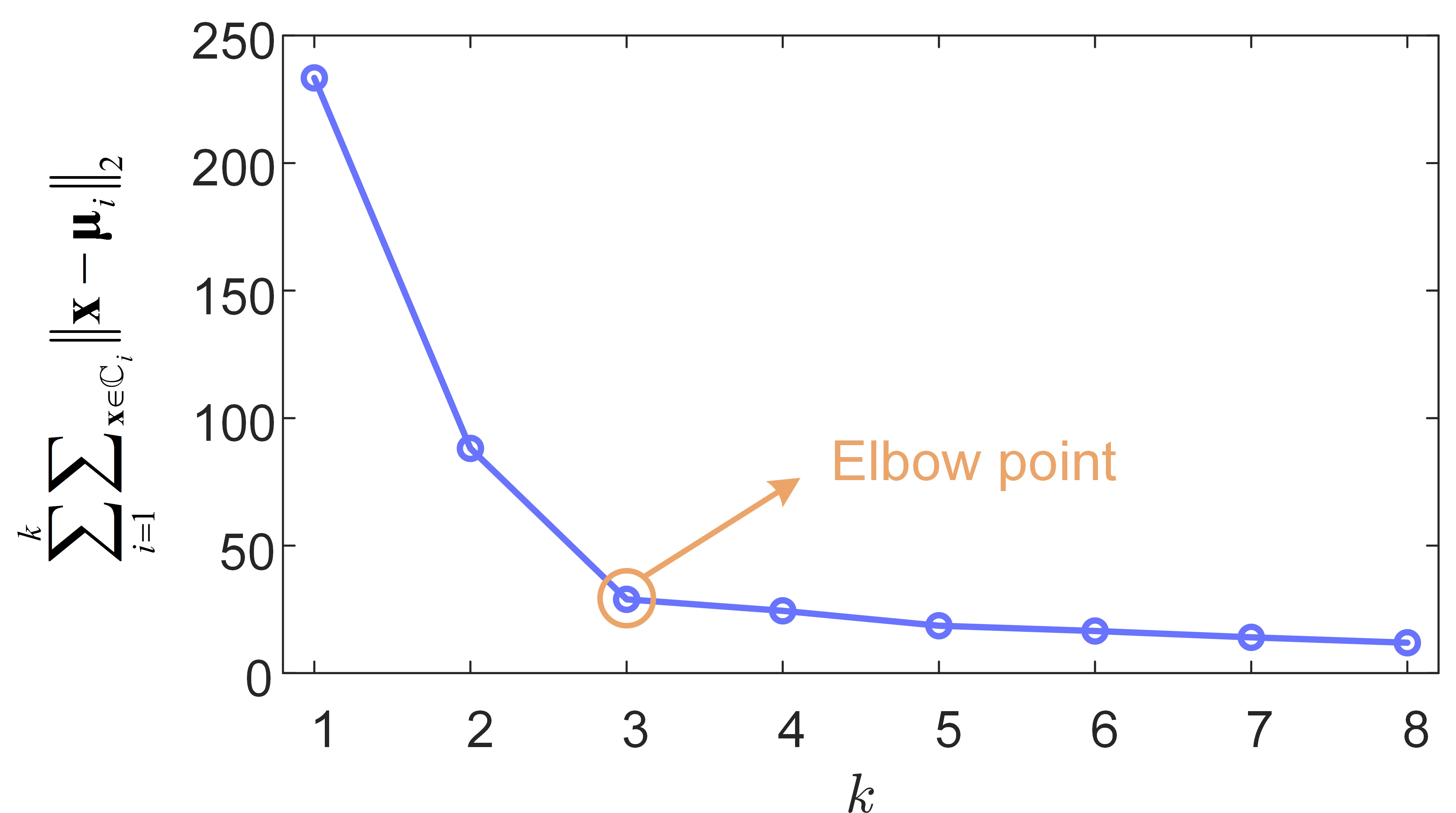}
\caption{The elbow point of the $k$-means method.}
\label{fig_2}
\end{figure}

\begin{figure*}[!t]
\centering
\includegraphics[width=6.5in]{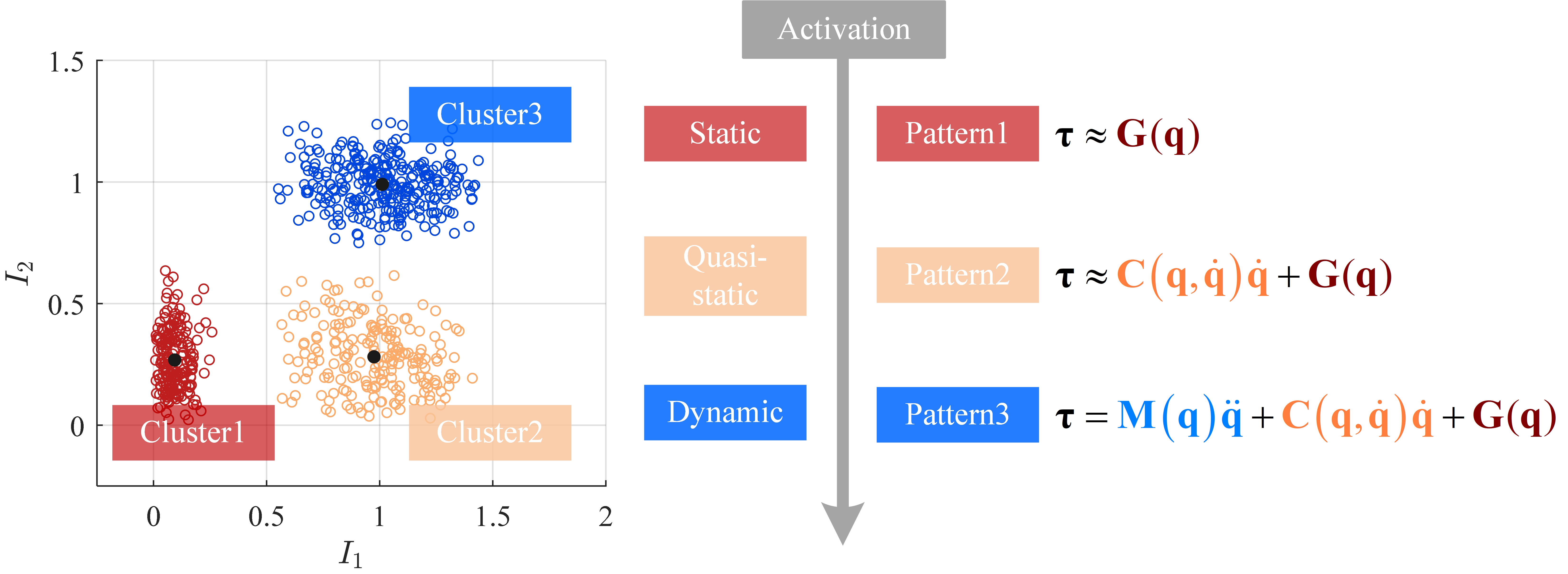}
\caption{The physical interpretation of three clusters.}
\label{fig_3}
\end{figure*}

For instance, we simulate the $n$-DoF serial manipulator and collect the motion signals under positioning and sorting working conditions. In the velocity and acceleration signals, we added 60 dB and 50 dB of white Gaussian noise, respectively. The diagram of data clustering for the collected motion signals according to the activation index is shown in Fig. 1. The results show that, after data processing by the activation index, the collected motion signals are explicitly classified into three clusters. It's worth noting that the number of clusters $k$ is selected by the elbow method\cite{ref17}. As shown in Fig. 2, the elbow point of $k$-means method based on activation index occurs at $k=3$.
\subsection{Physical interpretation of data clustering patterns}
In this subsection, we try to find the physical interpretation for the data clustering patterns. Tracing the data from the three clusters to the motion signals of the $n$-DoF manipulator, the three clusters correspond to the static, uniform velocity, and variable velocity motion states, respectively. Returning to the general dynamic equation of robotic manipulators, i.e., Eq. (\ref{eq_1}), the physical interpretation of the clustering patterns with the prior knowledge of dynamics is shown in Fig. 3. According to the prior knowledge, the motion signals in static state, quasi-static, and dynamic state sequentially activate the gravity term, the Coriolis and centrifugal terms, and the inertia items of the dynamical system, respectively.

For the dynamics model reconstruction task of robotic manipulators, stepwise reconstruction from the simplest form of steady-state motion signals can effectively reduce the complexity of symbolic regression. The reconstruction results of current step can be extended to the next step. Furthermore, this approach can effectively avoid the effect of velocity or acceleration noise on the reconstruction results of the current step. The stepwise strategy provides the potential to reconstruct the dynamics model of robotic manipulators based on SINDy method.

\section{Stepwise model reconstruction for ideal model}
According to the inclusion or exclusion of strong nonlinearities such as joint friction, the dynamics models of the $n$-DoF serial manipulators are classified as real or ideal models. This paper demonstrates the procedure and effect of two types of dynamics models reconstruction through simulation and experiment, respectively.

\subsection{Ideal model observation}
Ignoring nonlinearities such as joint friction, the ideal dynamics model of the $n$-DoF serial manipulator with all revolute joints satisfies the following properties.

\textbf{Property 1}. For the ideal dynamics model of the n-DoF serial manipulator, the joint torques can be written in the linear form of the inertia parameters 
\begin{equation}
\label{eq_8}
{\mathbf{\tau }} = {\mathbf{\psi }}\left( {{\mathbf{q}},{\mathbf{\dot q}},{\mathbf{\ddot q}}} \right){\mathbf{\theta }},
\end{equation}
where $\mathbf{\psi}$ and $\mathbf{\theta}$ respectively are the state quantity matrix and the inertia parameter vector of the $n$-DoF serial manipulator. Furthermore, Eq. (\ref{eq_1}) can be decomposed according to the result of data clustering in Section. 2 as
\begin{equation}
\label{eq_9}
{\mathbf{\tau }} = \underbrace {{{\mathbf{\psi }}_3}\left( {{\mathbf{q}},{\mathbf{\ddot q}}} \right){{\mathbf{\theta }}_3}}_{{\mathbf{M}}\left( {\mathbf{q}} \right){\mathbf{\ddot q}}} + \underbrace {{{\mathbf{\psi }}_2}\left( {{\mathbf{q}},{\mathbf{\dot q}}} \right){{\mathbf{\theta }}_2}}_{{\mathbf{C}}\left( {{\mathbf{q}},{\mathbf{\dot q}}} \right){\mathbf{\dot q}}} + \underbrace {{{\mathbf{\psi }}_1}\left( {\mathbf{q}} \right){{\mathbf{\theta }}_1}}_{{\mathbf{G}}\left( {\mathbf{q}} \right)}.
\end{equation}
The idea of stepwise model reconstruction based on SINDy is to replace the state quantity matrices with the libraries of candidate functions,  which can be expressed as
\begin{equation}
\label{eq_10}
{{\mathbf{\psi }}_i}{{\mathbf{\theta }}_i} \to {{\mathbf{Y}}_i}{{\mathbf{\Xi }}_i},{\text{ }}i = 1,2,3.
\end{equation}
With this, the reconstructed dynamics model of the $n$-DoF serial manipulator can be obtained through sparse regression.

\textbf{Property 2}. Since the $n$-DoF serial manipulator is all revolute joints, constant, trigonometric functions on angles, angular velocity and acceleration of joints are chosen as the elementary functions. Due to the presence of geometric nonlinearities in the $n$-DoF serial manipulator, the terms in the model with respect to the trigonometric functions on angles have a more complex form than the velocity and acceleration terms. Therefore, we introduce two feature sets \cite{ref12} to construct the library  candidate functions 
\begin{equation}
\label{eq_11}
\mathcal{F}_1 = \left\{ {\mathop \prod \limits_{i = 1}^{N_1} {c_{1i}}} \right\},
\end{equation}
where
\begin{equation*}
{c_{1i}} \in \left\{ 1 \right\} \cup \left\{ {\sin {q_1},\cos {q_1}, \cdots ,\sin {q_n},\cos {q_n}} \right\},
\end{equation*}
and
\begin{equation}
\label{eq_12}
\mathcal{F}_2 = \left\{ {\mathop \prod \limits_{i = 1}^{N_2} {c_{2i}}} \right\},
\end{equation}
where
\begin{equation*}
{c_{2i}} \in \left\{ {{{\dot q}_1}, \cdots ,{{\dot q}_n}} \right\} \cup \left\{ {{{\ddot q}_1}, \cdots ,{{\ddot q}_n}} \right\}.
\end{equation*}
$N_1$ and $N_2$ are the orders of the elementary functions and positively related to the complexity of the features. The complete feature set has the following form
\begin{equation}
\label{eq_13}
\mathcal{F} = {\mathcal{F}_1}{\mathcal{F}_2}.
\end{equation}
Each element in $\mathbf{\tau}$ can be obtained by a linear combination of the features in $\mathcal{F}$, therefore non-repeating features in $\mathcal{F}$ make up the library of candidate functions.
\subsection{Stepwise model reconstruction}
In this paper, a 3-DoF serial manipulator is introduced as the benchmark platform, because the first three axis in a classical serial manipulator determine the position of the end-effector in space and are more influenced by dynamics (e.g., gravity, inertia). The simulation of the 3-DoF serial manipulator under different motion states is completed with MATLAB-Robotic toolbox. The diagram of the 3-DoF serial manipulator is shown in Fig. 3.
\begin{figure}[!t]
\centering
\includegraphics[width=3in]{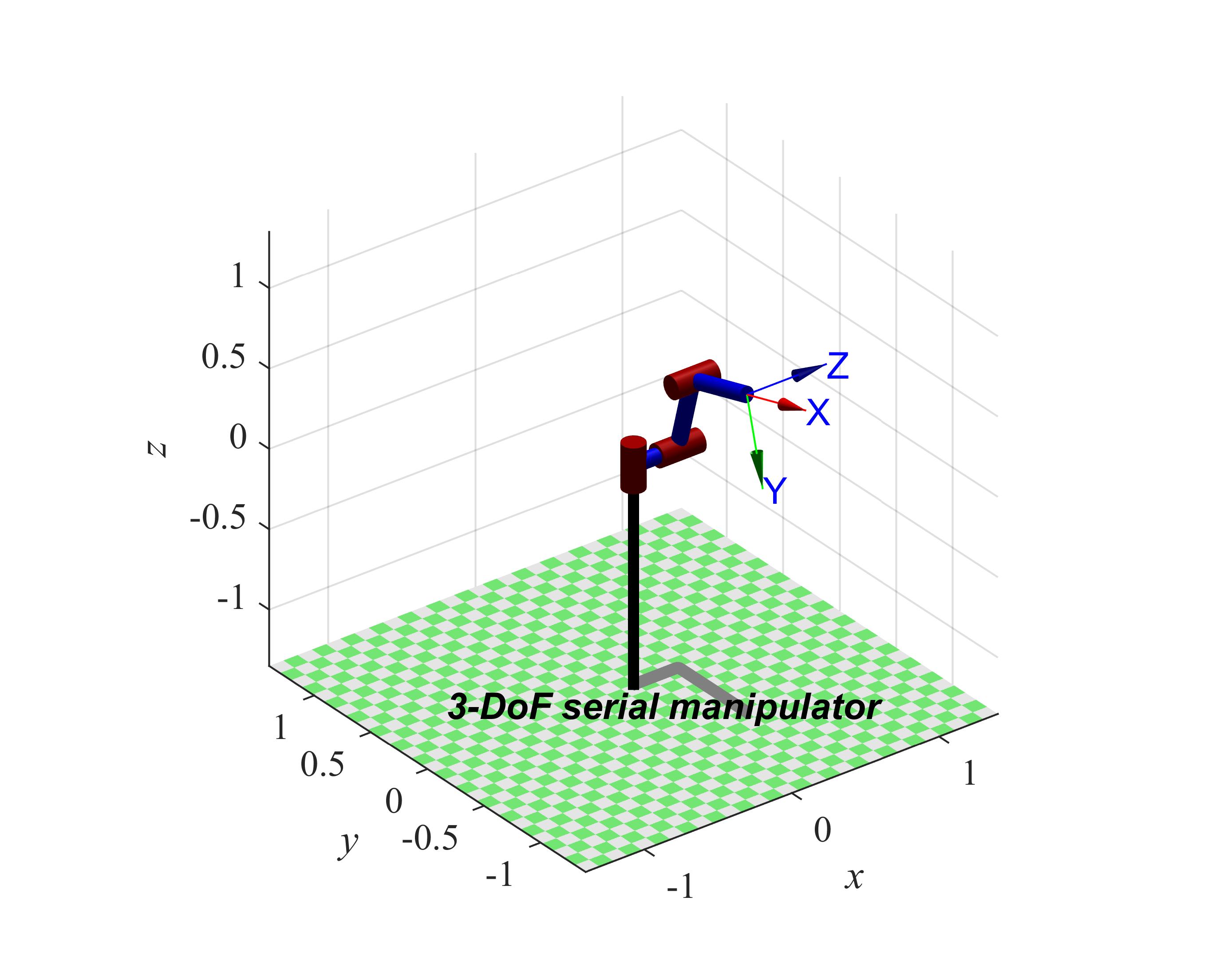}
\caption{The diagram of 3-DoF serial manipulator.}
\label{fig_3}
\end{figure}
Based on \textbf{Property 1}, the dynamic model the 3-DoF serial manipulator can be reconstructed in following three steps:

(1) Step1

In pattern1, the lowest activation data is only relevant to three joint angles, implying that the analytical expression form of the physical model behind these data is the most concise. Therefore, we choose the data under pattern1 for the first step of model reconstruction. For this case, the feature set $\mathcal{F}_2^{{S_1}}$ can be specifically chosen as
\begin{equation}
\label{eq_14}
\mathcal{F}_2^{{S_1}} = \left\{ 1 \right\},
\end{equation}

(2) Step2

In pattern2, the collected data is related to angles and angular velocities of the three joints. Therefore, the angular velocities are introduced into the elementary functions, and the feature set $\mathcal{F}_2^{{S_2}}$ can be specifically chosen as
\begin{equation}
\label{eq_15}
\mathcal{F}_2^{{S_2}} = \left\{ {\mathop \prod \limits_{i = 1}^{N_2^{{S_2}}} c_{2i}^{{S_2}}} \right\},
\end{equation}
where
\begin{equation*}
c_{2i}^{{S_2}} \in \left\{ {{{\dot q}_1},{{\dot q}_2},{{\dot q}_3}} \right\}.
\end{equation*}

(3) Step3

In pattern3, the collected data has the highest activation and is consistent with all state quantities. Therefore, the angular accelerations of the three joints are introduced into the elementary functions, and the feature set $\mathcal{F}_2^{{S_3}}$ can be specifically chosen as
\begin{equation}
\label{eq_16}
\mathcal{F}_2^{{S_3}} = \left\{ {\mathop \prod \limits_{i = 1}^{N_2^{{S_3}}} c_{2i}^{{S_3}}} \right\},
\end{equation}
where
\begin{equation*}
c_{2i}^{{S_3}} \in \left\{ {{{\dot q}_1},{{\dot q}_2},{{\dot q}_3}} \right\} \cup \left\{ {{{\ddot q}_1},{{\ddot q}_2},{{\ddot q}_3}} \right\}.
\end{equation*}

It is worth mentioning that the determination of the orders of the elemental functions in each feature set is crucial. For that, we collect the data separately and calculate the regression errors of the constructed library of candidate functions at different orders by the least squares method. As the number of orders increases, the error of least squares regression decreases rapidly after reaching a certain value, while the error does not decrease significantly when the number of orders continues to increase, implying that we have found the optimal number of orders. The regression errors of SINDy at different orders are shown in Fig. 5(a), that of stepwise SINDy at different orders are shown in Fig. 5(b), (c), (d), respectively. Fig. 5(e) is the regression error bar and the blue boxes in the figure indicate the optimal orders of SINDy and stepwise SINDy.
\begin{figure}[!t]
\centering
\includegraphics[width=3in]{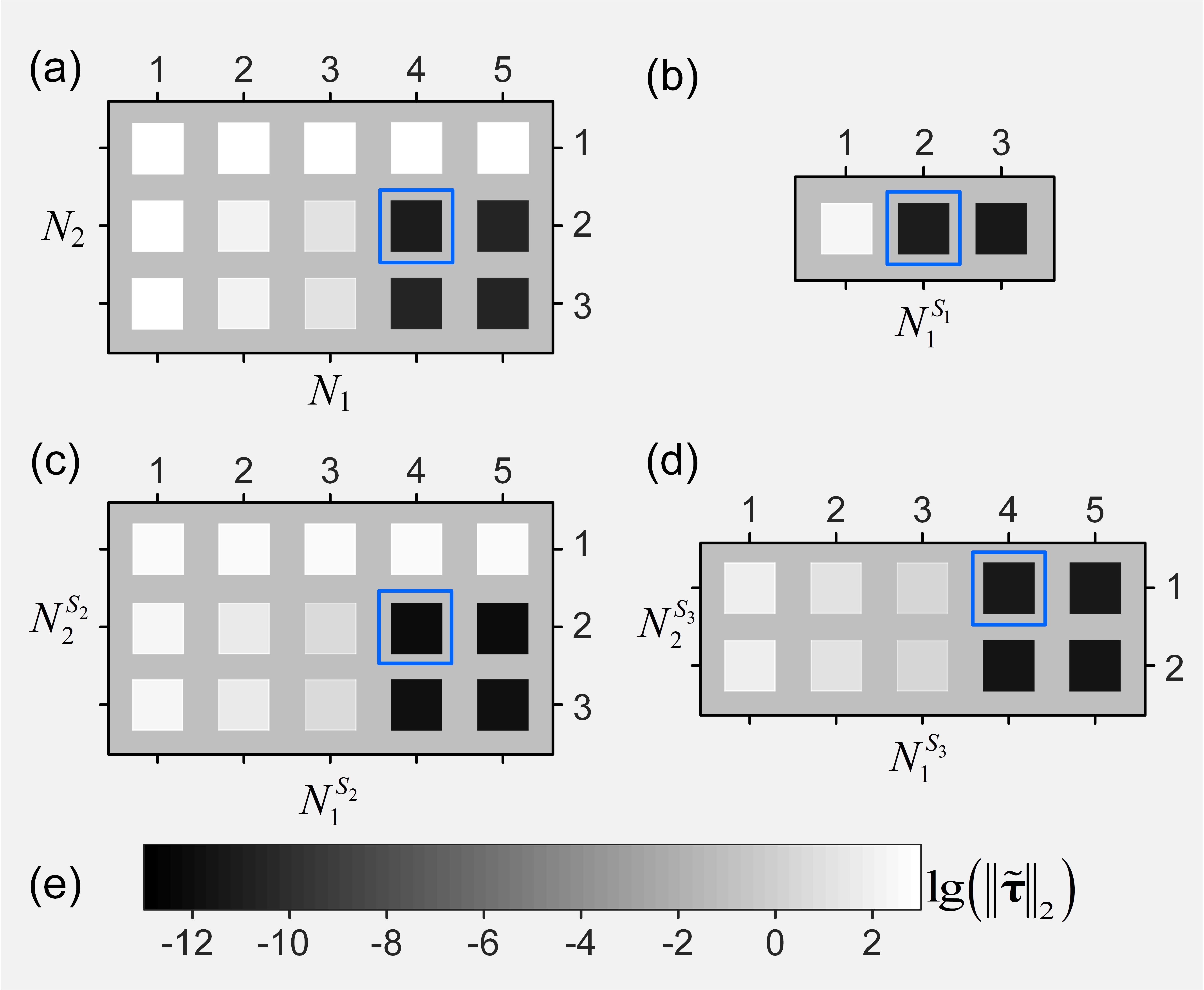}
\caption{The optimal orders of SINDy and stepwise SINDy.}
\label{fig_5}
\end{figure}
\begin{figure}[!t]
\centering
\includegraphics[width=3.5in]{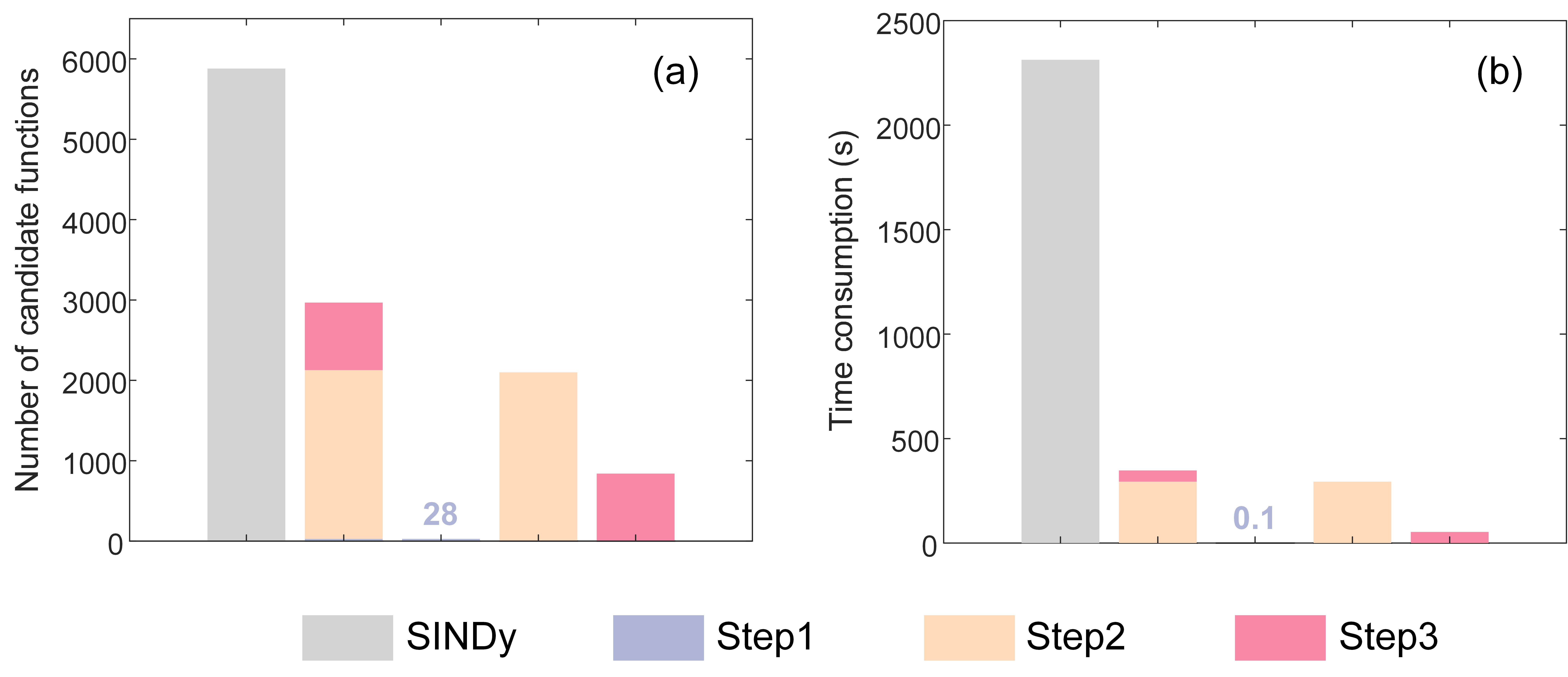}
\caption{Comparisons of SINDy and stepwise SINDy. (a) The number of candidate functions. (b) The computational efficiency.}
\label{fig_6}
\end{figure}

\subsection{Simulation Results}
Herein, the model of robotic manipulator is decomposed by $k$-means clustering method, and the construction method of library function is designed for each step separately. In comparison with the traditional SINDy, the SINDy-based stepwise model reconstruction method reduces the size of the library of candidate functions while significantly improving the computational efficiency of the regression problem. The comparison results are presented in Fig. 6. 

In addition, we also focus on the performance of the SINDy-based stepwise model reconstruction method considering data noise, and discuss the following three cases. In case1, all collected data is noise-free. In case2, the position, velocity and acceleration signals in the collected data possess 70 dB, 50 dB and 30 dB of noise, respectively. In case3, the collected data has more powerful noise, i.e., 60 dB, 40 dB, and 30 dB in the position, velocity, and acceleration signals, respectively. The terms in the library of candidate functions with an absolute value of the regression coefficient greater than 0.1 are defined as active terms. As shown in Fig. 7, the number of activation terms in the stepwise SINDy regression results is basically the same as that of the traditional SINDy in case 1, while the former is significantly lower than the latter in cases considering noise, i.e., case1 and case2. Fewer active terms means that the regression results are more sparse, as well as tending to be positively correlated with the extrapolation ability of the reconstructed model. In order to compare the extrapolation ability of the reconstructed models of the two methods considering noise, data beyond the training set is used for torque prediction. The torque prediction results for the extrapolation dataset are shown in Fig. 8. The results show that the reconstructed model using the stepwise SINDy still has a good extrapolation ability in case2 and case3, whereas the reconstructed model  using the traditional SINDy has failed to accurately predict the torques in case 3.

\begin{figure}[!t]
\centering
\includegraphics[width=3.25in]{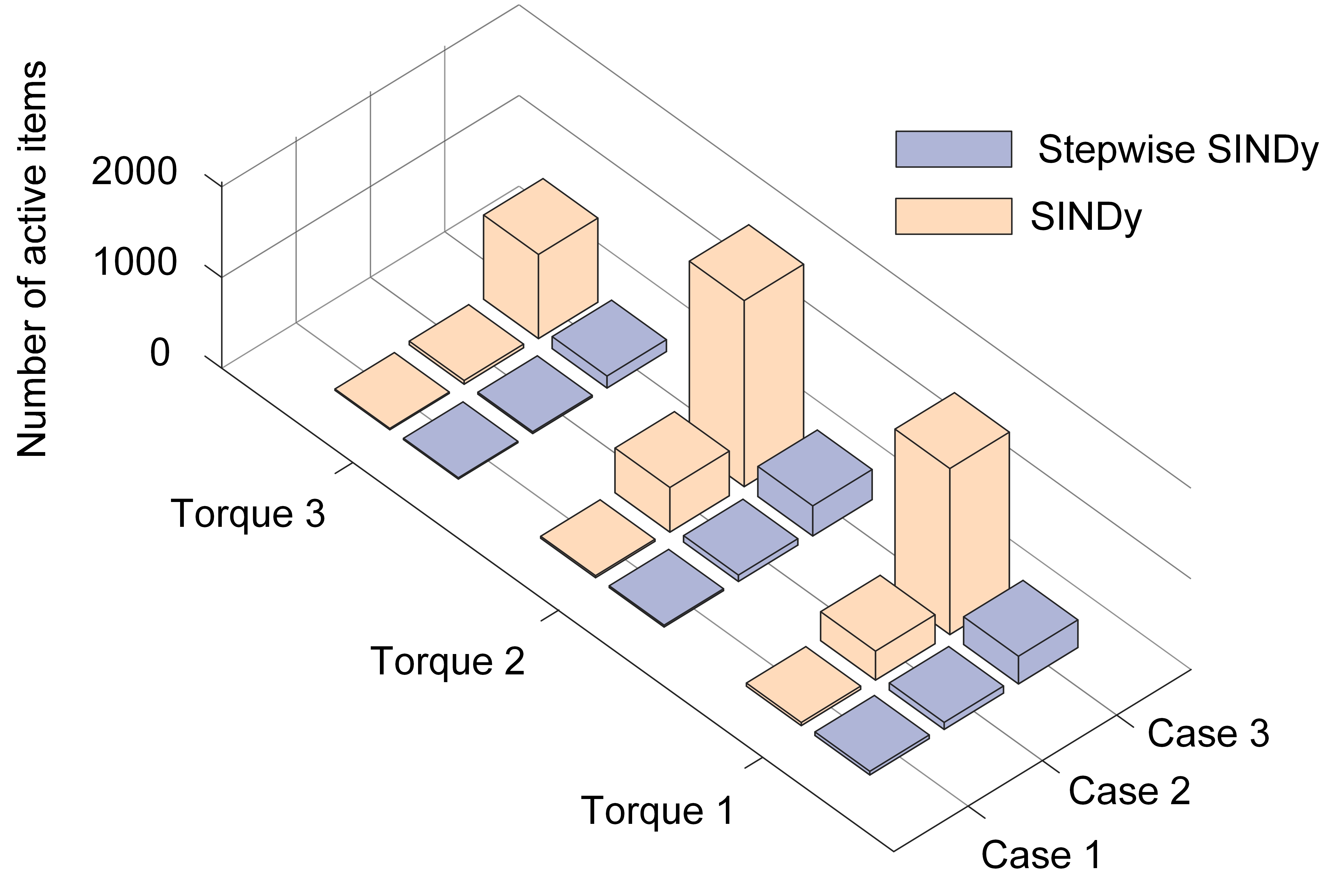}
\caption{The number of activation terms in regression results of SINDy and stepwise SINDy.}
\label{fig_7}
\end{figure}

\begin{figure}[!t]
\centering
\includegraphics[width=3.25in]{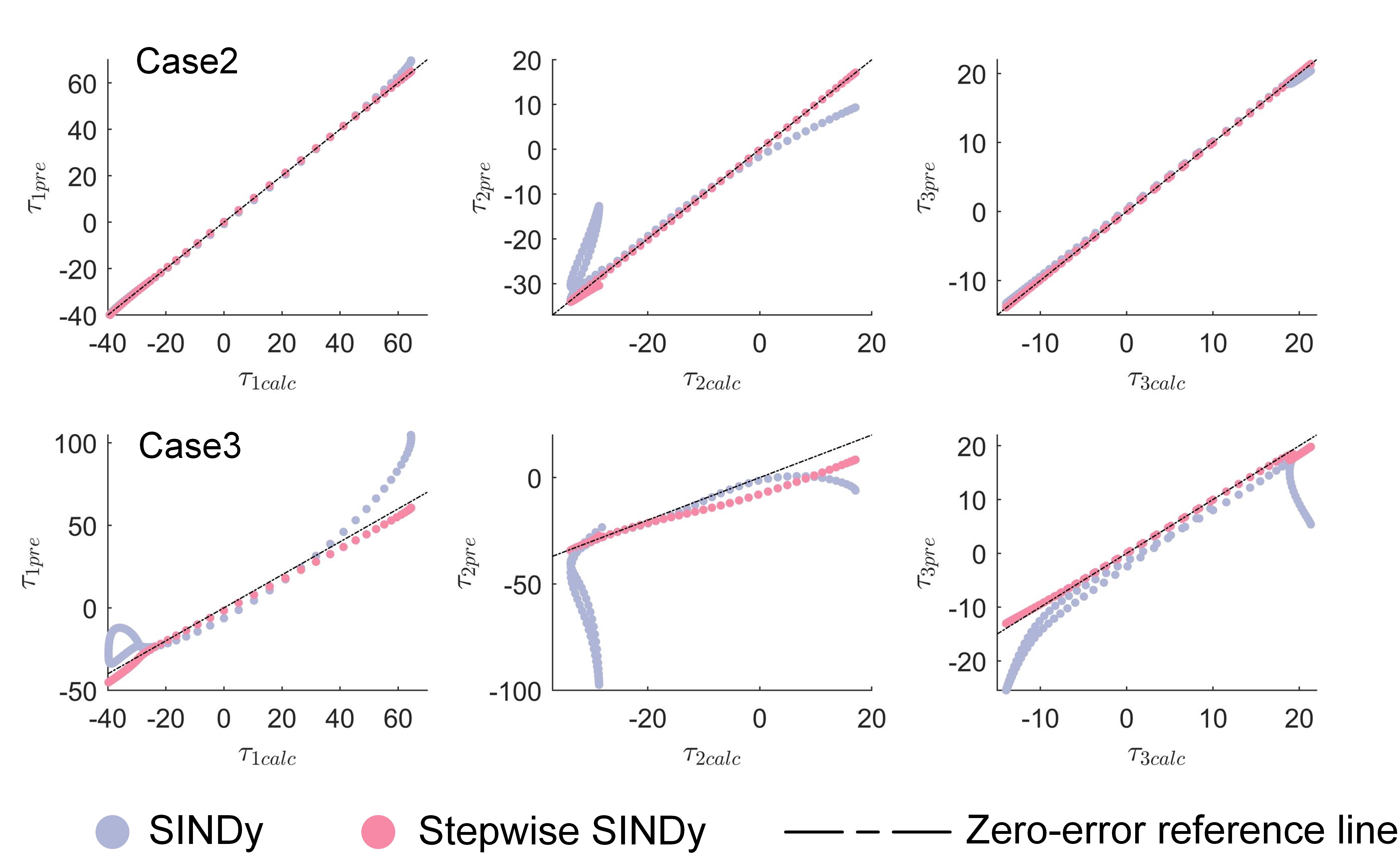}
\caption{Comparison of torque predictions between SINDy and stepwise SINDy using data external to the training set.}
\label{fig_8}
\end{figure}

\section{Experiment}
To validate the effectiveness of the SINDy-based stepwise model reconstruction method, we built an experimental platform of three-axis serial manipulator. The experimental platform is given in Fig. 9. The three-axis serial manipulator is equipped with three elephant joint modules and the controller is chosen as the speedgoat. In the experiment, we collect data in the speed mode of three joint modules and test the trajectory tracking performance of different control laws in the torque mode of the three joint modules with a control frequency of 1000Hz. 
\begin{figure}[!t]
\centering
\includegraphics[width=2.5in]{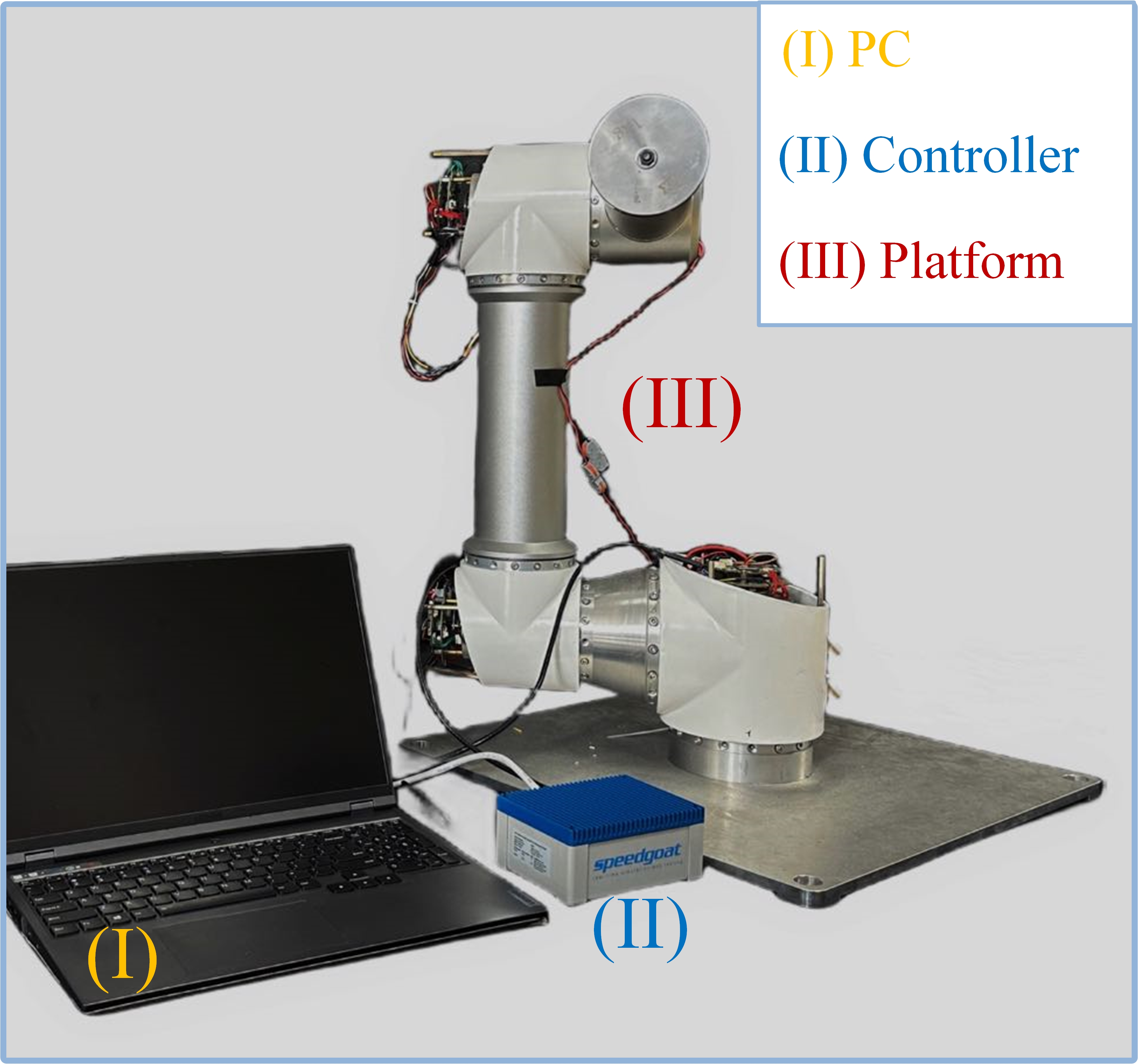}
\caption{The experimental setup of the three-axis serial manipulator.}
\label{fig_9}
\end{figure}
\begin{figure}[!t]
\centering
\includegraphics[width=3.2in]{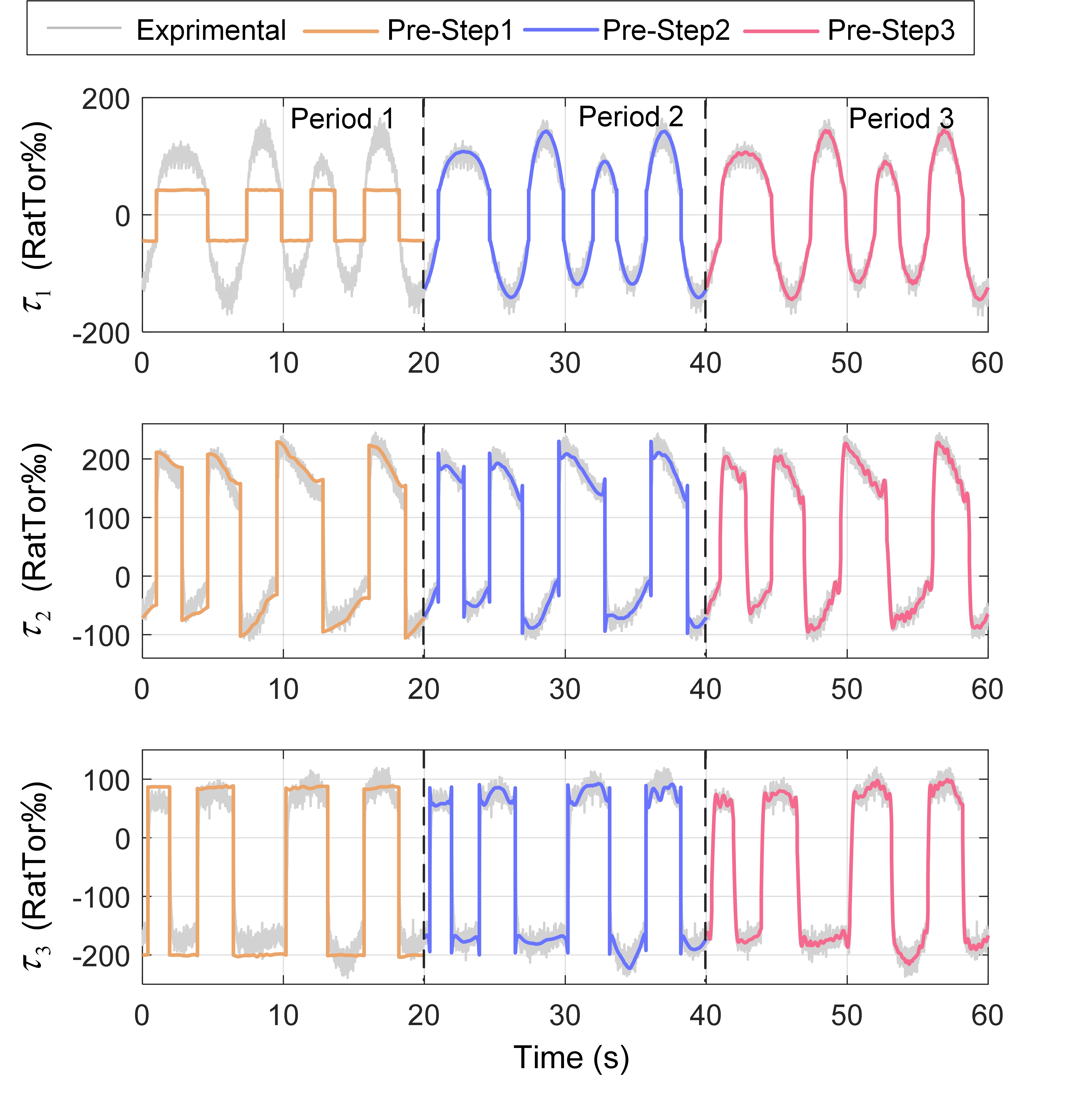}
\caption{Torque predictions of reconstructed models in different steps.}
\label{fig_10}
\end{figure}

For the experimental platform, there are nonlinearities in the model such as joint friction that cannot be ignored. Taking the classical stribeck friction model as an example
\begin{equation}
\label{eq_17}
f\left( v \right) = \left[ {{f_c} + \left( {{f_s} - {f_c}} \right){e^{ - {{\left( {{\raise0.7ex\hbox{$v$} \!\mathord{\left/
 {\vphantom {v {{v_s}}}}\right.\kern-\nulldelimiterspace}
\!\lower0.7ex\hbox{${{v_s}}$}}} \right)}^2}}}} \right]{\text{sign}}\left( v \right) + {f_v}v,
\end{equation}
there are two functions that cannot realize  parametric linearization, i.e., sign function and exponential function. For the sign function, we divide the collected data into two parts, forward and backward, and perform stepwise SINDy for each part of the data. For the exponential function, we add exponential functions with different coefficients to library of candidate functions, so as to increase the fitting ability of the library of candidate functions to strongly nonlinear factors. It is worth mentioning that during the step1 of data acquisition, we make the three joint modules perform uniform motion at very low speeds due to the need to overcome dry friction, so that the state of the system is approximate to static.

\begin{figure}[!t]
\centering
\includegraphics[width=3.2in]{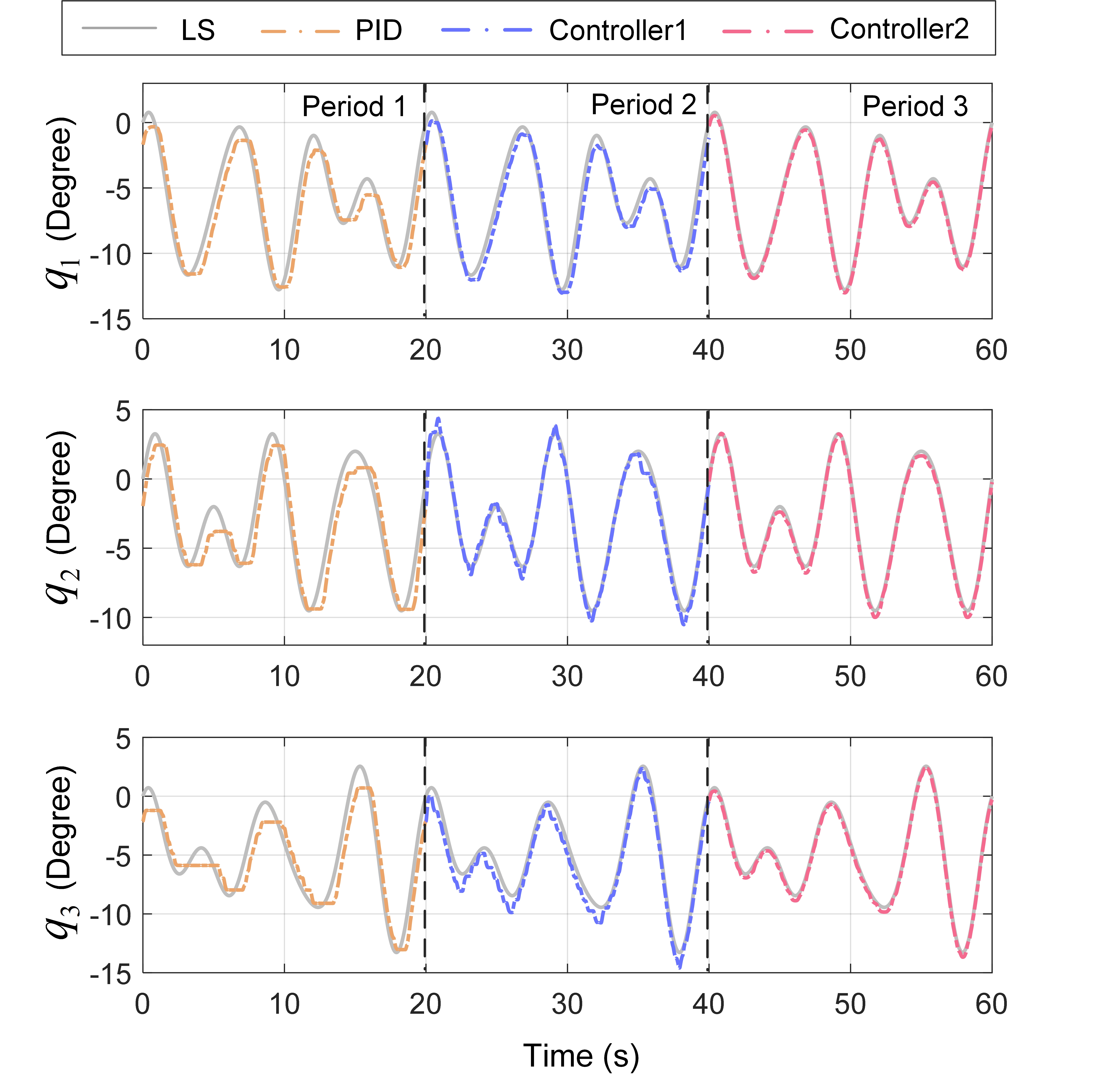}
\caption{The experimental setup of the three-axis serial manipulator.}
\label{fig_11}
\end{figure}

After three steps of data collection and subsequent stepwise SINDy, the reconstructed model corresponding to each step can be obtained. The stepwise reconstructed model becomes more and more complex, and consequently it becomes increasingly predictive. Fig. 10 demonstrates the ability of the three different reconstruction models to predict the torques for the same periodic trajectory, and the results show that the accuracy of the reconstructed models in predicting the joint torques of the three-axis serial manipulator becomes more accurate as the number of steps increases. Additionally, we deploy the reconstructed models to the controller and design the corresponding control strategies for the reconstructed models of step1 and step3, respectively. The control strategy for the model reconstructed in step1 is designed as
\begin{equation}
\label{eq_18}
{\mathbf{\tau }} = {K_p}{\mathbf{e}} + {K_i}\int_0^t {{\mathbf{e}}dT}  + {K_d}{\mathbf{\dot e}} + {\mathbf{\hat G}},
\end{equation}
where $\mathbf{\hat G}$ is obtained from the desired signal. The control strategy for the model reconstructed in step3 is designed as
\begin{equation}
\label{eq_19}
{\mathbf{\tau }} = {K_p}{\mathbf{e}} + {K_i}\int_0^t {{\mathbf{e}}dT}  + {K_d}{\mathbf{\dot e}} + {\mathbf{\hat M}}{{\mathbf{\ddot q}}_d} + {\mathbf{\hat C}}{{\mathbf{\dot q}}_d} + {\mathbf{\hat G}}.
\end{equation}
where the model compensation is also obtained from the desired signal. Traditional model-free control, i.e., PID control, is introduced as a contrasting. The trajectory tracking effect of the three controllers is shown in Fig. 11. In contrast to the model-free PID control, the controller1 that uses the reconstructed model in step1 as the dynamics compensation then has a significant improvement in trajectory tracking accuracy. The trajectory tracking accuracy is further improved by using the reconstructed model in step3 as the dynamic-compensated controller2. In practice, the compensation model can be selected for the accuracy and real-time performance of the controller.

\section{Conclusion}
This study contributed to provide a reliable data-driven model reconstruction method for multi-DoF robotic manipulators. Firstly, the limitations of the traditional SINDy method in dealing with the multi-freedom robotic arm problem are analysed, and a data clustering method using the activation index is designed to achieve the division of the dynamic model. Then, based on the clustering results, a stepwise model reconstruction method, i.e. stepwise SINDy, is proposed. The construction of library of candidate functions at each step in the stepwise SINDy is completely presented, and the advantages of the proposed method in terms of computational efficiency, sparsity, and extrapolation capability are verified by the 3-DoF simulation platform. Finally, the experimental platform of three-axis serial manipulator is established to validate the effectiveness of the SINDy-based stepwise model reconstruction method data-driven method. Experimental results show that the proposed method can accurately predict the joint torques and the excellent trajectory tracking performance of reconstructed control-based controllers also illustrate the applicability of the proposed method. Meanwhile, the interpretability of the reconstructed models in three steps also provides a basis for the subsequent design of more complex control laws.

There are two perspectives that need further refinement in the future work. Firstly, we will try to explore the possibilities of data mining algorithms for the dynamic modeling of robotic manipulators. For instance, using data noise and data correlation as indicators for unsupervised learning provides new ideas for dynamic modeling based on data-driven methods. Secondly, the model reconstruction for industrial 6-DoF robotic manipulator is still facing the problem of dimension curse, and we will attempt to deal with the high-dimensional system problem in a more concise and effective way.

\section*{Acknowledgments}
This work was supported by the National Natural Science Foundation of China under Grant No. 12072237.

\end{document}